\documentclass[conference]{IEEEtran}
\IEEEoverridecommandlockouts
\usepackage{cite}
\usepackage{amsmath,amssymb,amsfonts}
\usepackage{algorithmic}
\usepackage{algorithm}
\usepackage{graphicx}
\usepackage{colortbl}
\usepackage{color}
\usepackage{bbding}
\usepackage{diagbox}
\usepackage{threeparttable}
\usepackage{multirow}
\usepackage{makecell}
\usepackage{textcomp}

\usepackage{xcolor}
\def\BibTeX{{\rm B\kern-.05em{\sc i\kern-.025em b}\kern-.08em
		T\kern-.1667em\lower.7ex\hbox{E}\kern-.125emX}}
\begin{document}
	
	\title{SCPMan: Shape Context and Prior Constrained Multi-scale Attention Network for Pancreatic Segmentation*\\
		%{\footnotesize \textsuperscript{*}Note: Sub-titles are not captured in Xplore and
			%should not be used}
		\thanks{Corresponding author: Linlin Shen and Song Wu.}
	}
	
	\author{\IEEEauthorblockN{1\textsuperscript{st} Leilei Zeng }
		\IEEEauthorblockA{\textit{College of Computer Science and} \\
			\textit{Software Engineering}\\
			\textit{Shenzhen University}\\
			Shenzhen, China \\
			zengleilei123@gmail.com}
		\and
		\IEEEauthorblockN{2\textsuperscript{nd} Xuechen Li}
		\IEEEauthorblockA{\textit{College of Computer Science and} \\
			\textit{Software Engineering}\\
			\textit{Shenzhen University}\\
			Shenzhen, China \\
			timlee@szu.edu.cn}
		\and
		\IEEEauthorblockN{3\textsuperscript{rd} Xinquan Yang}
		\IEEEauthorblockA{\textit{College of Computer Science and} \\
			\textit{Software Engineering}\\
			\textit{Shenzhen University}\\
			Shenzhen, China \\
			xinquanyang99@gmail.com}
		\and
		\IEEEauthorblockN{4\textsuperscript{th} Linlin Shen}
		\IEEEauthorblockA{\textit{College of Computer Science and} \\
			\textit{Software Engineering}\\
			\textit{Shenzhen University}\\
			Shenzhen, China \\
			llshen@szu.edu.cn}
		\and
		\IEEEauthorblockN{5\textsuperscript{th} Song Wu}
		\IEEEauthorblockA{\textit{South China Hospital} \\
			\textit{Medical School}\\
			\textit{Shenzhen University}\\
			Shenzhen, China \\
			wusong@szu.edu.cn}
		\and
		%\IEEEauthorblockN{6\textsuperscript{th} Given Name Surname}
		%\IEEEauthorblockA{\textit{dept. name of organization (of Aff.)} \\
			%\textit{name of organization (of Aff.)}\\
			%City, Country \\
			%email address or ORCID}
	}
	
	\maketitle
	
	\begin{abstract}
		%Pancreatic cancer has a poor prognosis, hence accurate early detection and segmentation are critical for improving patient outcomes. However, pancreatic segmentation is challenged by blurred boundaries, high shape variability, and class imbalance. To tackle these problems, we propose a hybrid multi-scale attention network with shape context memory and ASM prior constraint for robust pancreas segmentation. Specifically, we proposed a Multi-scale Feature Extraction Module (MFE) and a Mixed-scale Attention Integration Module (MAI) to address unclear pancreas boundaries. Furthermore, a Shape Context Memory (SCM) module is introduced to jointly model semantics across scales and pancreatic shape and location. Experiments on NIH and MSD datasets demonstrate the efficacy of our model, advancing the state-of-the-art Dice Score for 1.01\% and 1.03\% respectively. Our architecture provides robust segmentation despite variations in pancreas shape, blurry boundaries, and scale.
		Due to the poor prognosis of Pancreatic cancer, accurate
		early detection and segmentation are critical for improving
		treatment outcomes. However, pancreatic segmentation is challenged
		by blurred boundaries, high shape variability, and class imbalance.
		To tackle these problems, we propose a multiscale
		attention network with shape context and prior constraint for robust pancreas segmentation. Specifically,
		we proposed a Multi-scale Feature Extraction Module (MFE) and
		a Mixed-scale Attention Integration Module (MAI) to address
		unclear pancreas boundaries. Furthermore, a Shape Context
		Memory (SCM) module is introduced to jointly model semantics
		across scales and pancreatic shape. Active Shape Model (ASM) is further used to modeo the shape priors. Experiments on
		NIH and MSD datasets demonstrate the efficacy of our model,
		which improve the state-of-the-art Dice Score for 1.01\% and 1.03\%
		respectively. Our architecture provides robust segmentation performance, against the blurry boundaries, and variations in scale and shape of pancreas.
	\end{abstract}
	
	\begin{IEEEkeywords}
		Medical Image Segmentation, Activate Shape Model
	\end{IEEEkeywords}
	
	\section{Introduction}
	Pancreatic cancer is one of the most lethal malignancies with an inferior prognosis and a 5-year survival rate of less than 9\%. Therefore accurate early detection and appropriate treatment play a critical role in improving its prognosis \cite{n1,n2}. Accurate segmentation of pancreatic images as a prerequisite for detecting pancreatic cancer, can effectively eliminate interferences of complex background and benefit physicians for accurately assessing disease progression and treatment response. Computed tomography (CT) can clearly show the structure of the pancreas and is the most widely used examination method for pancreatic cancer detection by clinical imaging physicians \cite{n3,n4,n5}. However, manually outlining the pancreas is very tedious, time-consuming, and unable to meet clinical needs. Therefore there is an urgent need for a system that can automate the segmentation of the pancreas, which can help imaging physicians to conduct early screening and quantitative detection of pancreatic lesions.
	
	Different from other abdominal organs, such as the liver \cite{n6}, lung \cite{n7}, and kidney \cite{n8},  which can be effectively segmented by AI-based system, the segmentation accuracy of the pancreas is not high enough to be applied to clinical practise, due to the following challenges: 1) Unclear boundaries. The boundary of the pancreas cannot be well defined due to the problem of fuzzy boundary perturbation caused by the similar density of pancreas and the surrounding tissues (e.g., stomach wall, duodenum, small intestine). 2) High variability in shape. The shape, size, and location of the pancreas varies greatly across individuals. This makes it difficult for artificial intelligence (AI) to represent its shape and location. 3) The pancreas occupies only a small portion of the entire CT image. There is a serious imbalance between size of the target and the background, which leads to overfitting problem on the background region.
	
	Given these challenges, traditional machine learning image segmentation algorithms struggle to accurately delineate pancreatic boundaries where the edges are unclear \cite{n9,n10}. The same problem happens in natural segmentation when the boundaries are blurred or ambiguous \cite{n11,n12}.
	% Previous works [14,15] utilize multi-scale feature information to enable scale analysis of objects. The issue of blurred object-background boundaries is often addressed by fusing features across scales to enable joint recognition. 
	Prior works \cite{n13,n14} exploit multi-scale features to address, blurred boundaries by fusing cross-scale representations.
	In pancreas segmentation, the combination of multi-scale features also achieves promising performance \cite{n15,n16}. However, pancreas segmentation presents challenges beyond boundary ambiguity. Unlike other abdominal organs (e.g. kidney, liver) with relatively fixed morphology, the pancreas exhibits high variability in shape across individuals. The irregular shape and complex morphology of the pancreas lead to difficulties and poor robustness in modeling its shape representations. Li et al. \cite{n17} proposed to use attentional mechanisms and multi-scale convolution to enrich contextual information and prevent the learning of irrelevant regions, which reduced the impact of inter-class similarity, and achieved 86.10\% Dice Score on the NIH dataset. Davradou.A \cite{n18} proposed to solve the pancreas segmentation problem by applying a modified U-Net model to the cropped data, based on a morphological active contour algorithm. However, its Dice score is relatively low, which is only 67.67\%. Yu et al. \cite{n19} proposed a recurrent saliency transformation network \cite{n20}, which allowed the network to be jointly optimized as well as propagate multi-stage visual information to improve segmentation accuracy.
	% Previous work has only focused on a single problem such as blurred pancreas boundaries or large variations in pancreas shape [16,17,18], while work such as [19] is based on a two-step approach where a coarse segmentation is performed first and the results of the coarse segmentation are used to first localize the pancreas position followed by a fine segmentation.
	Boundary ambiguity and high shape variability, which are intrinsic characteristics of the pancreas, are main challenges for localization and segmentation. However, previous approaches either ignore or fail to effectively model these properties. Prior arts tackling shape variation or boundary ambiguity demonstrate limited performance \cite{n15,n16,n17}, and two-step approach e.g. like \cite{n18} is more complex and resource-consuming.
	
	In this work, we propose a hybrid multi-scale attention shape context memory and an ASM (Activate Shape Model) prior constraint based pancreas segmentation network, to address both blurred pancreas boundaries and large shape variations problems. The main contributions of this is work are:
	\begin{itemize}
		\item A Multi-scale Feature Extraction Module (MFE) and a  Mixed-scale Attention Integration Module (MAI) are introduced to pancreas segmentation to tackle the issues of blurred pancreas boundaries.
		
		\item A shape context memory (SCM) module is proposed to jointly model pancreas location and shape representation.
		
		\item An ASM prior constraint module is developed to guide pancreas shape modeling to provide shape priors.
		
		\item A weighted binary cross-entropy (BCE) and an Iou loss function are used to address the imbalance problem between the size of background and pancreas.
	\end{itemize}
	\section{Related work}
	
	In the previous subsections, the significance and difficulties of pancreas segmentation studies were summarized and presented. In the next subsections, the process of pancreas segmentation studies is reviewed in detail.
	The rapid development of deep learning has led to an explosion of techniques for pancreas segmentation in abdominal CT scans. This segmentation can be divided into traditional machine learning methods and deep learning-based methods.
	\subsection{Traditional machine learning methods}
	In traditional machine learning, heuristic methods, such as the approach developed by Tam et al. \cite{n45}, involve selecting seed points within the pancreas and using region growing techniques for segmentation.Shan et al. \cite{n46} segment the pancreas by distinguishing it from surrounding tissues using Otsu \cite{n47} threshold algorithms and morphological operations. However, these methods rely heavily on the selection of threshold ranges and are often challenged by subtle density differences between the pancreas and adjacent tissues, leading to potential inaccuracies. To address these limitations, researchers including Erdt \cite{n48} and Karasawa \cite{n49} et al. have developed more effective graph-based algorithms, although these algorithms still introduce considerable human generalization bias and rely on manual feature intervention, limiting their applicability for automated segmentation.
	\subsection{Deep learning based methods}
	In the field of deep learning, the success of these techniques in natural image processing has stimulated their application in medical image analysis \cite{n50,n51,n52}. Deep learning-based pancreas segmentation aims to mitigate the generalization bias inherent in traditional methods. These methods are mainly divided into two categories: convolutional neural network (CNN)-based and transformer-based methods. Take Zhou et al. \cite{n52} as an example, a CNN-based approach employs a full convolutional network (FCN) fixed-point model to improve iterative accuracy in small organ segmentation.Li et al. \cite{n20} proposed an advanced pancreas segmentation model combining a dual adversarial network and a pyramid pool module, achieving a Dice coefficient of 83.31 ± 6.32\%. Furthermore, Li et al. \cite{n38} enhanced the U-Net model using a multi-scale attention-dense residual U-net (MAD-UNet), resulting in more accurate feature acquisition and a Dice coefficient of 86.10 ± 3.52\% on the NIH dataset.Yu et al. \cite{n17} proposed a multi-stage saliency segmentation approach, using the segmentation mask of the current stage to improve the input of subsequent stages. However, the limited receiving range of CNNs hinders their focus on distant regions and to some extent ignores rich global context information, making it difficult to further optimize network performance. To overcome these limitations, transformer-based approaches have been developed.Qiu et al. \cite{57} introduced residual transformer UNet (RTUNet) for pancreas segmentation, achieving a Dice similarity coefficient of 86.25 ± 4.52\% on the NIH dataset. Dai et al. \cite{45} proposed a two-stage trans-deformer network, merging 2D Unet, deformable convolution, and wavelet-based multi-input module, showing excellent performance on both the NIH and MSD datasets. In addition, Zhu et al. \cite{58} recognized the 3D nature of CT data and explored the effectiveness of 3D networks, achieving a dice coefficient of 84.59 ± 4.86\% on the NIH dataset.
	Zhang et al. \cite{n19} developed a segmentation framework that combined multi-spectral registration with 3D level set techniques to refine probabilistic graphs to improve pancreatic edge prediction accuracy, achieving a dice coefficient of 84.47 ± 4.36\% on the NIH dataset.
	However, Transform-based approaches usually require a two-stage approach because initial localization of the pancreatic region is required first, while the extensive GPU memory requirements of 3D networks usually lead to CT scans being segmented into smaller portions or processed at reduced resolution, limiting spatial context learning.
	Therefore, developing a network that can effectively utilize the context relationship of pancreatic CT sequences in a single stage is crucial for accurate pancreatic segmentation.
	\section{METHODS}
	\begin{figure*}[htbp]
		\centerline{\includegraphics[width=0.95\textwidth]{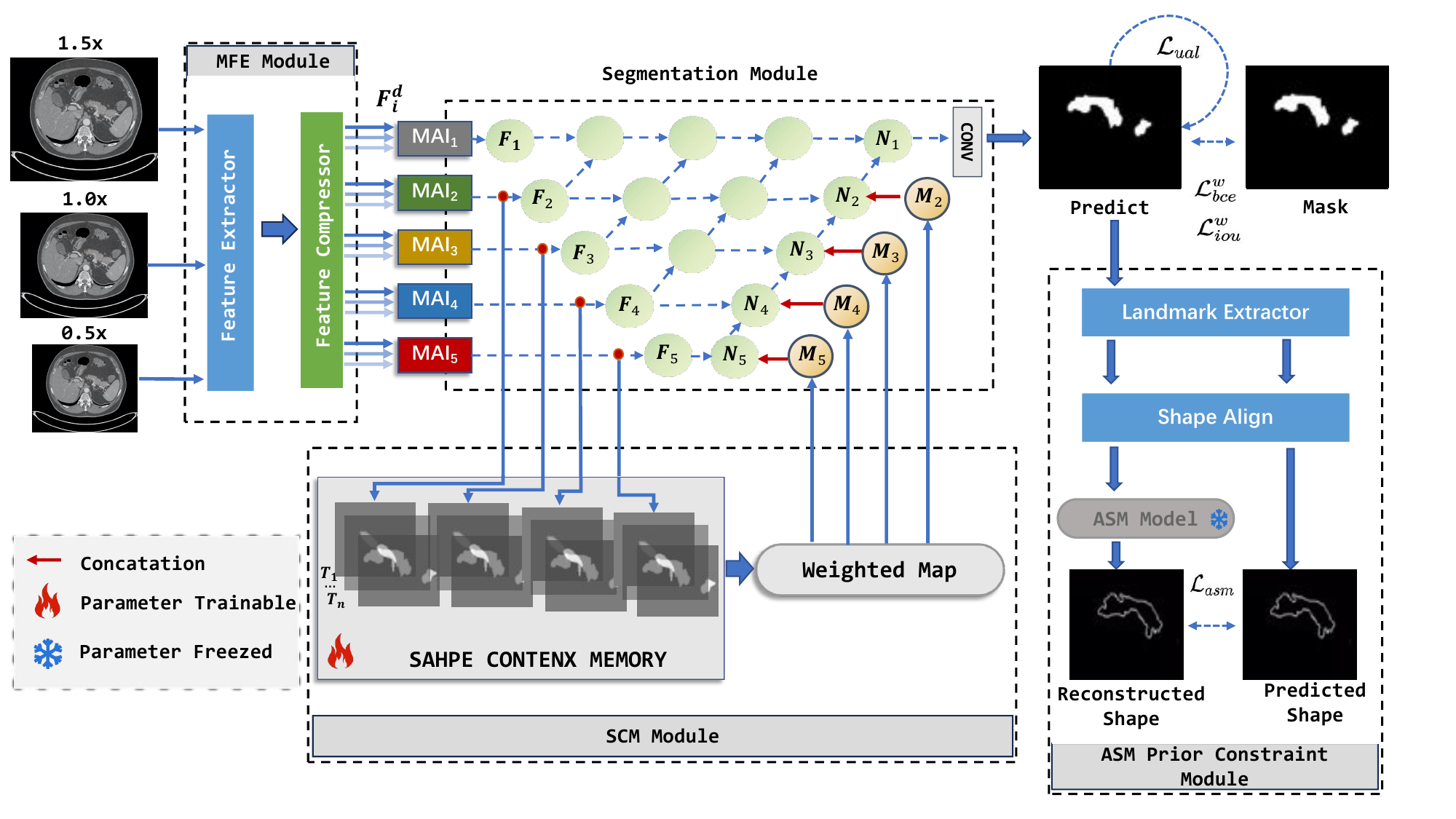}}
		\caption{The architecture of our proposed network. Firstly, a Multi-scale Feature Extractor (MFE) module is designed to extract the multi-scale feature ($F_{i}^{d}$). Successively, a Mixed-scale Attention Integration (MAI) module is designed to integrate the multi-scale feature $F_{i}^{d}$ to 1.0x ($F_{i}$). More, Shape Context Memory (SCM) module is designed to enhance the network's ability to model the shape and location of the pancreas and generate the enhanced feature $M_{i}$. Finally, the concatenated feature $N_{i} = [F_{i}, M_{i}], i\in[2,3,4,5]$ are fed to the Unet++ decoder for mask generation.}
		\label{fig1}
	\end{figure*}
In this work, a shape context and prior constrained multi-scale attention Network is proposed for the key challenges in pancreas segmentation i.e., blurred boundary, various shape and unbalanced foreground and background. As shown in Fig.~\ref{fig1} for blurred boundaries, we design a Multi-scale Feature Extraction (MFE) module and a Mixed-scale Attention Integration (MAI) module, inspired by camouflage detection \cite{n13}. For shape modeling, we incorporate anatomical priors through a Shape Context Memory (SCM) module to implicitly guide location and shape prediction, as well as an ASM Prior Constraint module for explicit shape regularization. Finally, we handle class imbalance using weighted BCE and IoU losses. 
% As shown in  The proposed SeqPrioShapeNet integrates the proposed MFE, MAI, HASM and ASM modules built upon a UNet++ backbone \cite{n21}.
\subsection{Multi-scale Feature Extraction Module.}\label{AA}
\begin{figure}[htbp]
	\centerline{\includegraphics[width=0.5\textwidth]{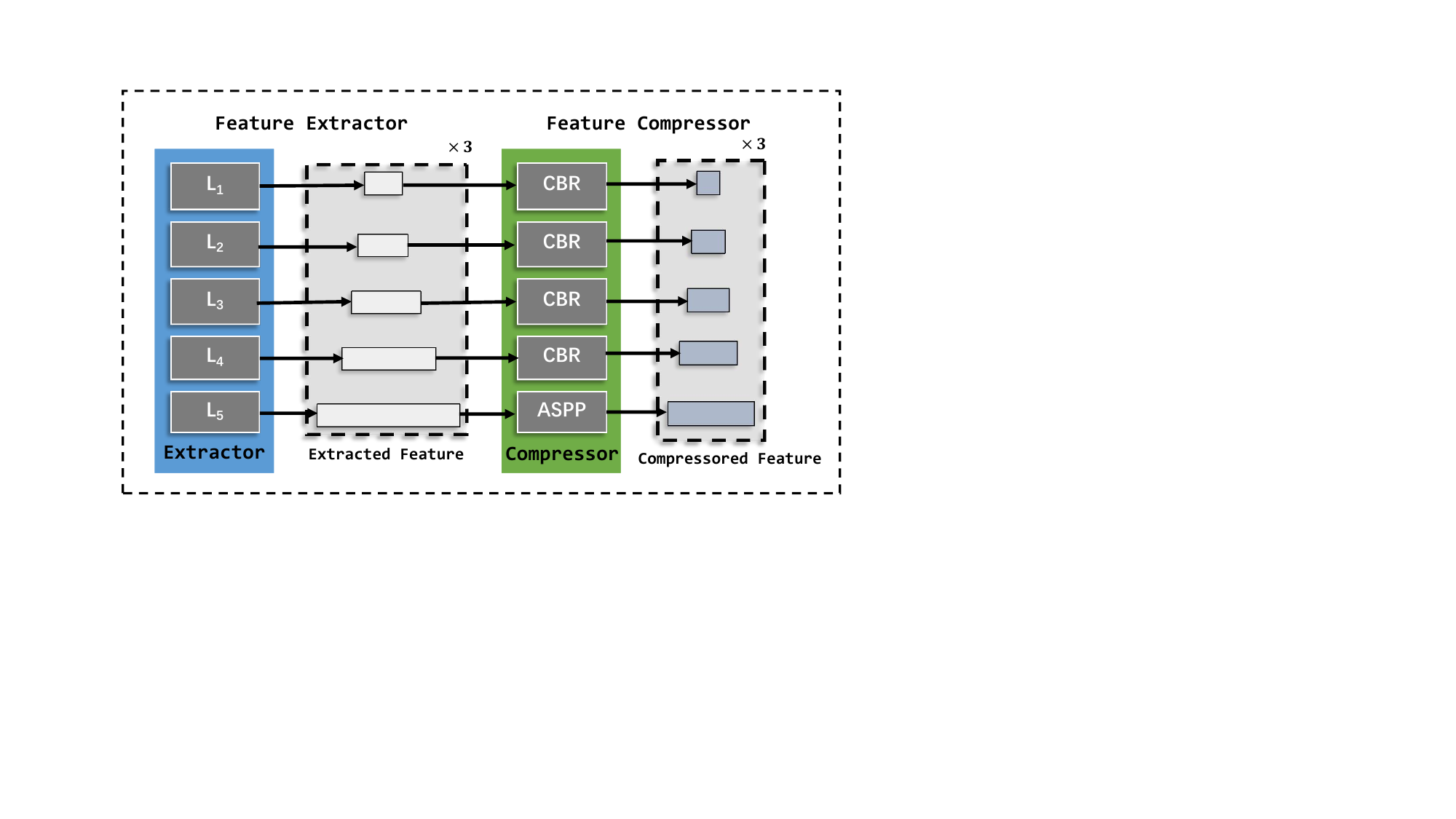}}
	\caption{Multi-scale Feature Extraction (MFE) Module. The MFE module consists of two main components: a feature extractor and a feature compressor, the feature extractor mainly uses the structure of Resnet-50, and the feature compressor uses CBRs for compression (the last layer of feature uses ASPP), and finally output the multi-scale features $F_{i}^{d}$. }
	\label{figMutil}
\end{figure}

The multi-scale feature extraction module consists of a feature extractor and a feature compressor (Fig.~\ref{figMutil}). 
% The feature extractor module uses the feature extraction encoder Resnet-50 [23] but removes the layers after the fourth layer. Specifically, the feature maps before passing through the first pooling layer are used and the output feature maps of "Layer1", "Layer2", "Layer3" and "Layer4" are used as inputs to the feature compression module to perform feature channel compression, in order to match the input of Unet++. The feature compressor module uses simplified ASPP[24] for feature compression extraction except for the input of the last layer, i.e., the input of the "Layer4", and the rest of the inputs are composed of modular units using independent CBR (include Conv-BatchNorm-Relu), with the optimized computation of the feature compression extraction module, we will get more compact features and finally, we will have a number of channels of 32, 64, 128, 256, 512.
The proposed feature extractor module adopts ResNet-50 \cite{n22} as the backbone, but removes the FC layer after the last ResNet block. We take the feature maps before the first max pooling as "$L_{1}$", term the remaining layers as "$L_{2}$" to "$L_{5}$" corresponding to ResNet blocks 1-4. To match UNet++ input size, the five extracted features are fed into feature compressor module, using simplified ASPP \cite{n23} for "$L_{5}$" and CBR (include Conv-BatchNorm-Relu) for "$L_{1-4}$".
\begin{equation}
	F^{d}_{i} = MFE(I^{d}),d\in[0.5x,1.0x,1.5x],i\in[1,2,3,4,5]\label{eqmutilencoder}
\end{equation}
where $I^{d}\in \mathbb{R}^{H \times W \times 3}$ is the input, including 3 different scales, $F^{d}_{i}$ is the output Multi-scale features.

\begin{figure}[htbp]
	\centerline{\includegraphics[width=0.5\textwidth]{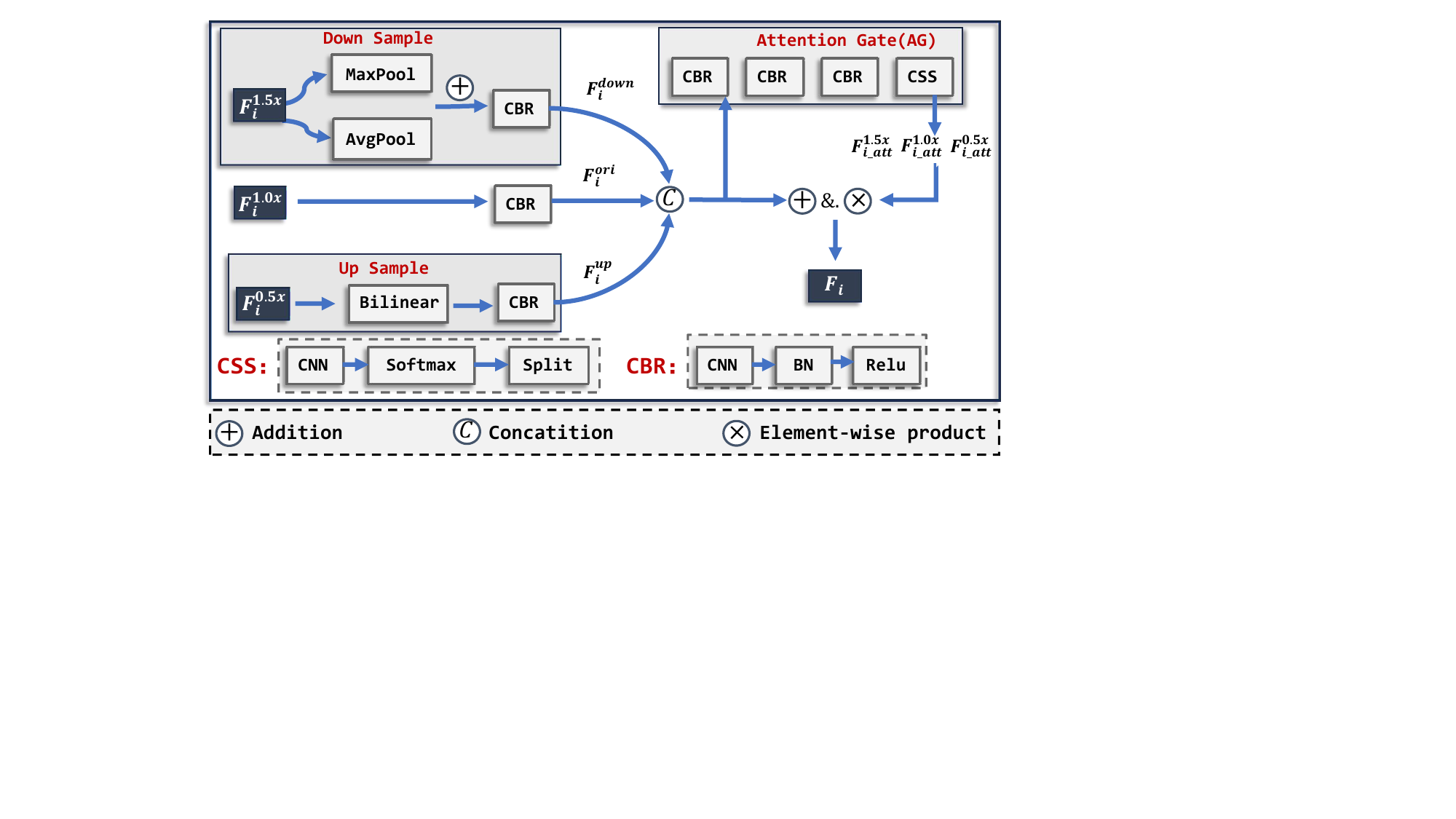}}
	\caption{Mixed-scale Attention Integration (MAI) Module. MAI module takes as input the MFE output features $F_{i}^{d}$, $i\in[1,2,3,4,5]$, $d\in[0.5\times,1.0\times,1.5\times]$ of the MFE, and generate the integrated multi-scale features $F_{i}$.}
	\label{fig2}
\end{figure}
\subsection{Mixed-scale Attention Integration Module.}\label{BB}
The scale-space theory, which uses image and feature pyramids to construct features integrating mulit-scales and semantics, has been widely and effectively applied in natural images to deepen the model's multi-scale structural understanding \cite{n24,n25,n26,n27}. Influenced by \cite{n13}, we use the Mixed-scale Attention Integration (MAI) module (e.g., Fig.~\ref{fig2}) to solve the problem of feature fusion at different scales.

Specifically, we take the feature $F_{i}^{d}$ extracted by MFE module and use MAI module to integrate $F_{i}^{d}$ to the 1.0x scale ($F_{i}$). The processing flow is shown in Algorithm 1.
% \begin{equation}
	% \begin{split}
		% &F_{i}^{down} = CBR([M(F_{i}^{1.5x})+A(F_{i}^{1.5x})],\theta_{1} ) \\
		% &F_{i}^{up} = CBR([B(F_{i}^{0.5x})],\theta_{2} ) \\
		% &F_{i}^{ori} = CBR(F_{i}^{1.0x},\theta_{3} ) \\
		% &F_{i} = Concat(F_{i}^{down},F_{i}^{ori},F_{i}^{up}) \\
		% &F^{0.5x}_{i\_att} ,F^{1.0x}_{i\_att},F^{1.5x}_{i\_att} = AG(F_{i}) \\
		% &F_{i} = F^{0.5x}_{i\_att} \times F^{0.5x}_{i} 
		%       + F^{1.0x}_{i\_att} \times F^{1.0x}_{i}  
		%       + F^{1.5x}_{i\_att} \times F^{1.5x}_{i} \label{eq1}
		% \end{split}
	% \end{equation}
\begin{algorithm}
	\renewcommand{\algorithmicrequire}{\textbf{Input:}}
	\renewcommand{\algorithmicensure}{\textbf{Output:}}
	\caption{MAI Processing}
	\begin{algorithmic}[1]
		\REQUIRE ~\
		$F_{i}^{d}$: The multi-scale feature from MFE module, $i\in[1,2,3,4,5], d\in[0.5\times,1.0\times,1.5\times]$ \
		\ENSURE ~\
		$F_{i}$: The integrated feature \
		
		\STATE $F_{i}^{down} = CBR([M(F_{i}^{1.5\times})+A(F_{i}^{1.5\times})])$
		
		\STATE $F_{i}^{up} = CBR([B(F_{i}^{0.5\times})])$
		
		\STATE$ F_{i}^{ori} = CBR(F_{i}^{1.0\times})$
		
		\STATE $F_{i} = Concat(F_{i}^{down},F_{i}^{ori},F_{i}^{up})$
		
		\STATE $F^{0.5\times}_{i\_att},F^{1.0\times}_{i\_att},F^{1.5\times}_{i\_att} = AG(F_{i})$
		
		\STATE $F_{i} = F^{0.5\times}_{i\_att} \times F^{0.5\times}_{i} 
		+ F^{1.0\times}_{i\_att} \times F^{1.0\times}_{i}  
		+ F^{1.5\times}_{i\_att} \times F^{1.5\times}_{i}$
		\STATE Return $F_{i}$
		% \ENSURE Compute MSE(S,R)
	\end{algorithmic}
\end{algorithm}
% For each FdiF_{i}^{d}, MM is MaxPool, AA is AvgPool, BB is Bilinear, we integrate its scale to 1.0x using the following MAI module. Where CBRCBR indicates the stacked “Conv-BN-ReLU” layers in the attention generator, and θ\theta means the parameters of these layers. ConcatConcat represents the concatenation operation.FdowniF_{i}^{down},FupiF_{i}^{up},ForiiF_{i}^{ori} represent the features obtained after DownsampleDownsample, UpsampleUpsample and CBRCBR only, respectively.

$F_{i}^{d}$ denotes the multi-scale features, where $M$ is max pooling, $A$ is average pooling, and $B$ is bilinear interpolation. We integrate the features to 1.0x scale using the proposed MAI module, which consists of downsample, upsample and CBR ("Conv-BN-ReLU") and CBS ("Conv-BN-Split") layers. Specifically, $F^{down}_{i}$, $F^{up}_{i}$ and $F^{ori}_{i}$ are the intermediate features after downsample, upsample and CBR operations, respectively. The final integrated features are computed by weighted sum of the attention-modulated multi-scale features after the AG operations.

\subsection{Shape Context Memory Module}\label{CC}
CT sequences exhibit continuity in organ shape and appearance across slices. Therefore, we propose a shape context memory (SCM) module to improve contextual feature learning. The high variability in pancreas shape, size and location makes big challenges for direct localization in individual slices. In contrast, our SCM module is designed to model contextual information of the entire image sequence, encoding the average spatial location and shape of the pancreas. This representation helps to guide accurate localization and shape modeling of the pancreas within the abdomen.
% Consecutive samples in CT sequences exhibit continuity, with organ shape and appearance correlated across slices. Motivated by this observation, we propose a hybrid multi-scale attention shape memory-guide module to cache pancreas sequence shape information and improve contextual feature learning. The high variability in pancreas shape, size and location makes direct localization challenging. In contrast, our HASM module is designed to model contextual information of the entire image sequence, encoding the average spatial location and shape of the pancreas. This representation helps to guide localization and deformation modeling of the pancreas within the abdomen.
\begin{figure}[htbp]
	\centerline{\includegraphics[width=0.5\textwidth]{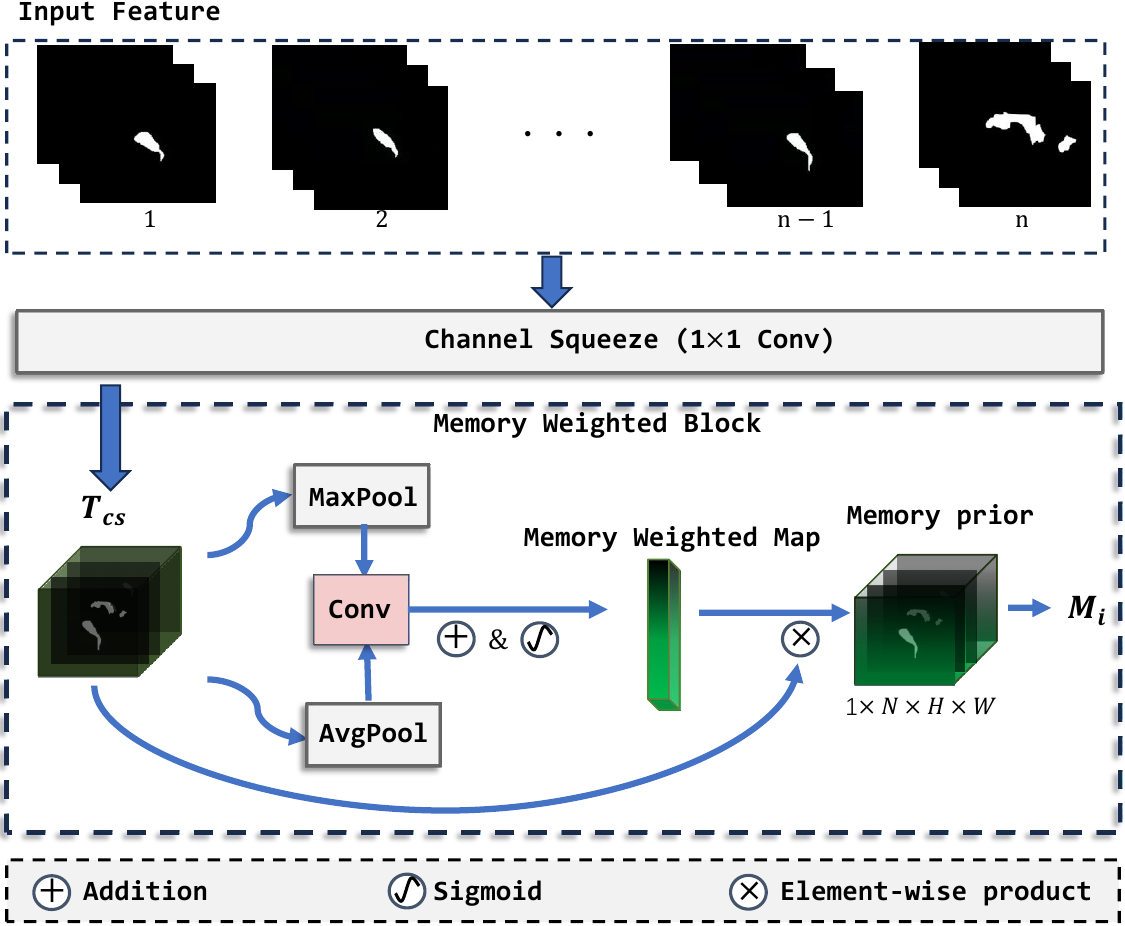}}
	\caption{Shape Context Memory (SCM) Module.}
	\label{fig3}
\end{figure}
% We first set the number of T slices to be cached at a time, we cache the features after the MAI module, and in order to improve the utilization of information, we cache only the features of the second to fifth layers after the output of the MAI module. And in order to minimize the unnecessary redundant information caused by constant caching, we weighted and scored the T-slices (e.g., Fig.~), and set higher weights for the caches that can improve the accuracy of the subsequent pancreas segmentation and take a higher proportion in the subsequent information fusion. Finally, we enter the decoder stage by fusing the filtered caches together with the current information fusion features. The process is formulated as:
We cache intermediate features from the outputs of MAI 2-5 with a batch of N slices. To minimize redundancy, the cached features are weighted by their contributions for subsequent pancreas segmentation, i.e. useful cached slices receive higher weights. The cached features are finally concatenated and fed into the decoder. This process can be formulated as:
\begin{equation}
	\begin{split}
		&T_{cs} = CS([T_{1},T_{2}...T_{n}]),T_{cs}\in \mathbb{R}^{1 \times N \times H \times W} \\
		&M_{i} = WB(T_{cs}), M_{i}\in \mathbb{R}^{1 \times N \times H \times W}\\
	\end{split}
\end{equation}
$WB$ and $CS$ denote the “Memory Weighted Block” and “Channel Squeeze” in Fig.~\ref{fig3}. 
\subsection{Segmentation Module}\label{bbone}
We adopt UNet++ \cite{n21} as segmentation module, with some modifsication in the encoding stage. Rather than downsampling to derive multi-scale features, we directly extract UNet++ input features at different scales from the output of the MAI module. In the decoding stage, we concatenate the shape context features $M_{i}$ learned from the SCM module with the output features $F_{i}$ of the MAI module to obtain $N_{i}$, which is then used for upsampling to derive the final segmentation result. The formal expression of this process is formula \ref{eqN}.
\begin{equation}
	\begin{split}
		&N_{i} = Concat(F_{i}, M_{i}), N_{i}\in \mathbb{R}^{N \times C \times H \times W}\label{eqN}
	\end{split}
\end{equation}
\begin{algorithm}
	\renewcommand{\algorithmicrequire}{\textbf{Input:}}
	\renewcommand{\algorithmicensure}{\textbf{Output:}}
	\renewcommand{\algorithmicreturn}{\textbf{Return:}}
	\caption{ASM Training}
	\begin{algorithmic}[1]
		\REQUIRE \
		% I=i1,i2,...,iNI = {i_1, i_2, ..., i_N}: Sample training images\\
		$I = {I_1, I_2, ..., I_N}$: Sample training images\\
		\ENSURE  \
		$\overline{s}$: Mean Shape\
		$P$: Shape Prior\
		$b$: Shape Deforms \
		\STATE Locate landmarks $s_{i}$ from $I_{i}$ using sobel and interpolation algorithm
		
		$s_{i} = [x_{1},y_{1},..,x_{n},y_{n}]$,
		$s_{i} \in\mathbb{R}^{2n\times 1}, n=238$
		%$s_{i}^{sobel}$ = $sobel(I_{i}), i \in [1, N]$, extract boundary using %sobel \
		
		%uniform $s_{i}^{sobel}$ length using 1-dimensional interpolation algorithm,\ 

		%$s_{i}$ = $interp1d(s_{i}^{sobel})$, 
		\STATE Align all landmarks $S = {s_1, s_2, ..., s_N}$
		\STATE Calculate the average shape $\overline{s} = \frac{1}{N}\sum_{i=1}^{N} s_i$
		\STATE Construct the shape difference matrix $D = S - \overline{s}$
		\STATE Build the covariance matrix: $C = D^{T}D$
		\STATE Use PCA (Principal Component Analysis) analysis to obtain shape prior $P$ and deform coefficients $b$\\
		$P,b \leftarrow PCA(C)$ \\
		\RETURN $\overline{s}$, $P$, $b$
		% \RETURN ¯s\overline{s}, PP, bb
	\end{algorithmic}
\end{algorithm}

\begin{algorithm}
	\renewcommand{\algorithmicrequire}{\textbf{Input:}}
	\renewcommand{\algorithmicensure}{\textbf{Output:}}
	\renewcommand{\algorithmicreturn}{\textbf{Return:}}
	\caption{ASM Based Shape Reconstruction}
	\begin{algorithmic}[1]
		\REQUIRE ~\
		$M$: Predicted mask \
		$\overline{s}$: Mean Shape \
		$P$: Shape Prior
		\ENSURE ~\
		
		$S^{\prime}$: The Reconstructed Shape 
		$S_0$: Shape of the Predicted Mask
		\STATE Choose the largest $area > threshold$ as $M_{lagest}$ from $M$ 
		\STATE Locate landmarks  $S_0$ from $M_{lagest}$ using sobel and interpolation algorithm
		
		$S_0 = [x_{1},y_{1},..,x_{n},y_{n}]$,
		$S_0 \in\mathbb{R}^{2n\times 1}, n=238$
		
		%$S_{0}^{sobel}$ = $sobel(M_{largest})$, extract boundary using sobel \
		
		%$S_{0}$ = $interp1d(S_{0}^{sobel})$, uniform $S_0^{sobel}$ scale using interp1d, 

		\STATE Align $S_0$ to $\overline{s}$ 
		
		\STATE Compute deform parameters $b$  \\
		$b=P^{T}\times (S_0-\overline{s}) $, $P \in \mathbb{R}^{2n \times 9}$, $b \in \mathbb{R}^{9 \times 1}$
		
		\STATE Reconstruct shape $S'$ \\
		$S^{\prime} = \overline{s} + Pb, S^{\prime} \in \mathbb{R}^{2n \times 1}$
		
		\RETURN $S^{\prime},S_0$
		% \ENSURE Compute MSE(S,R)
	\end{algorithmic}
\end{algorithm}

\begin{algorithm}
	\renewcommand{\algorithmicrequire}{\textbf{Input:}}
	\renewcommand{\algorithmicensure}{\textbf{Output:}}
	\renewcommand{\algorithmicreturn}{\textbf{Return:}}
	\caption{$L_{asm}$ Calculation}
	\begin{algorithmic}[1]
		\REQUIRE ~\
		
		$M$: Predicted mask \
		$R$: Shape Reconstruction (Algorithm 3) \
		$\overline{s}$: Mean Shape \
		$P$: Shape Prior
		\ENSURE ~\
		
		$\mathcal{L}_{asm}$: The ASM Loss
		
		% \IF{n>5n > 5}
		
		\STATE Generate reconstructed shape $S'$ and predicted mask shape $S_0$ \\
		$S^{\prime}, S_0 \leftarrow R(M, \overline{s}, P)$ \\
		
		\STATE $\mathcal{L}_{asm}=MSE(S', S_0) $
		
		\RETURN $\mathcal{L}_{asm}$
		% \ENSURE Compute MSE(S,R)
	\end{algorithmic}
\end{algorithm}
\subsection{ASM Prior Constraint module}\label{FF}
We further improve pancreas shape modeling by incorporating anatomical shape priors through ASM constraints. Specifically, we firstly construct an ASM model with a shape prior constraints based on representative pancreatic shape features. The ASM model is trained as described in Algorithm 2, and its output mean shape $\overline{s}$ and shape prior $P$ are used by Algorithm 3 to reconstruct the shape of predicted mask. When applying the trained ASM model for shape regularization in our proposed network, thresholding is performed on the predicted pancreas masks to remove small region to avoid the problems that prediction maps contain multiple regions. The largest region with size greater than the threshold is selected. Based on typical pancreas size, the threshold is set as 50 pixels. The resulting $\mathcal{L}_{asm}$ term, calculated as the $MSE$ between the reconstructed shape ($S^\prime$) and predicted mask ($S_0$) output by Algorithm 3, is incorporated into the final loss. The approach is summarized in Algorithm 4.

\subsection{Loss Formulation}\label{loss}
In the proposed model, the total training loss consists of four sub-losses:
\begin{equation}
	\mathcal{L}_{total} = \mathcal{L}^{w}_{Iou} + \mathcal{L}^{w}_{BCE} + \mu \times \mathcal{L}_{ual} + \tau \times \mathcal{L}_{asm}\label{eq6}
\end{equation}
where $\mathcal{L}^{w}_{Iou}$ and $\mathcal{L}^{w}_{BCE}$ represent the weighted IoU loss and binary cross entropy (BCE) loss, respectively, which have been widely adopted in segmentation tasks to imbalanced forground and background problems. We use the same definitions as \cite{n28,n29,n30}, since their effectiveness has been validated in these works. In this work,  $\mathcal{L}_{ual}$ \cite{n13} is employed to further optimize the segmentation for blurred edges that are difficult to define, which is formulted as $\mathcal{L}_{ual}^{i,j} = 1 - \left |2M_{i,j} - 1 \right |^{2}$ and the $M_{i,j}$ denote the predicted value at position $(i,j)$. Where $\mu$ is the $\mathcal{L}_{ual}$ balance coefficient. Following previous work \cite{n13}, we use the best equilibrium coefficient strategy cosine as the equilibrium coefficient strategy in this paper. The $\mathcal{L}_{asm}$ is the ASM prior constraints loss and $\tau$ is the balance coefficient.
% \begin{equation}
	% \begin{split}
		% &N_{i} = Concat(F_{i}, M_{i}), N_{i}\in \mathbb{R}^{N \times (C+1) \times H \times W}\label{equUAL}
		% \end{split}
	% \end{equation}
\section{EXPERIMENTS}
\begin{figure}[htbp]
	\centerline{\includegraphics[width=0.5\textwidth]{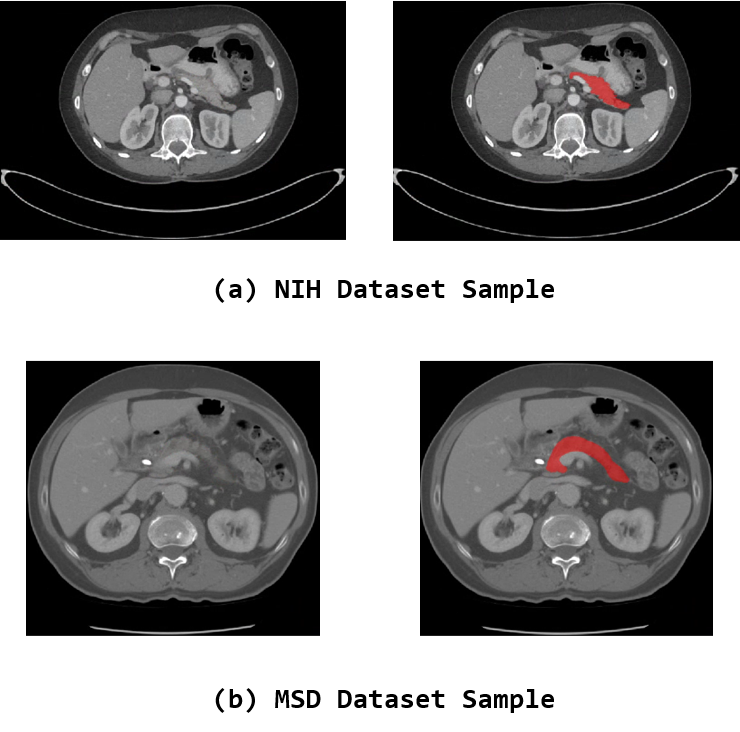}}
	\caption{Example images in NIH dataset and MSD dataset.}
	\label{fig4}
\end{figure}
\subsection{Datasets and Evaluation Metrics}\label{DD}
\textbf{Datasets.} We conducted experiments on two publicly available pancreas segmentation datasets: 1) NIH dataset: which consists of 82 abdominal contrast-enhanced CT scans from the National Institutes of Health (NIH) Clinical Center pancreas segmentation dataset \cite{n31}, 2) MSD dataset: which consists of 281 labeled pancreas and pancreatic tumors from the Medical Segmentation Decathlon (MSD) Challenge pancreas segmentation dataset \cite{n5}. Examples of images from these two datasets with their corresponding labels are visualized in Fig.~\ref{fig4}.

For each CT Volume in the NIH dataset, the volume size is $\left [512\times   512 \times D \right ] $, where $D\in \left [181,466  \right ]$ is the number of slices along the transverse plane. The slice thickness of the CT scans varies from 1.5 mm to 2.5 mm, depending on the scan depth. We follow the principle of four-fold cross-validation and randomly divide the NIH dataset into four fixed subsets , which include 21, 21, 20, and 20 samples, respectively.

For each CT Volume in the MSD dataset, the volume size is $\left [512\times   512 \times D \right ] $, where $D\in \left [37,751  \right ]$ is the number of slices along the transverse plane. Following previous work [34] on the MSD dataset, we combine the pancreas and pancreatic tumor into a single entity as the segmentation target. We divide the dataset into four subsets containing 70, 70, 70, and 71 CT volumes respectively, for cross-validation.

\textbf{Data preprocessing.} In medical imaging, especially in computed tomography (CT) scans, choosing the best range of CT values is crucial to obtain clear and accurate images. For the specific case of pancreatic imaging, as a rule of thumb, we limit CT values to the range  as a means of preconditioning. This specific range was identified as significantly enhancing the contrast in areas associated with the pancreas.

\textbf{Evaluation Metrics.} Dice score, Precision, and Recall are employed as the evaluation criteria. The definitions of the above three evaluation criteria are as follows:
\begin{equation}
	\begin{split}
		Precision = \frac{TP}{TP + FP}\label{eq3}
	\end{split}
\end{equation}
\begin{equation}
	\begin{split}
		Dice = \frac{2TP}{2TP + FP + FN}\label{eq4}
	\end{split}
\end{equation}
\begin{equation}
	\begin{split}
		Recall = \frac{TP}{TP + FN}\label{eq5}
	\end{split}
\end{equation}
Here TP, FP and FN represent Ture positive, False positive and False Negative.
% \textbf{Baseline Network architectures.} For the multi-scale feature extraction module, following recent works [35,36,37,38], we initialize the feature extractor module using pre-trained ResNet-50 parameters on ImageNet, while the rest of the network is randomly initialized. For the segmentation module, we use Unet++, an end-to-end segmentation network with excellent performance in medical image segmentation, as our backbone.
\subsection{Implementation Details}\label{IMPD}
SGD with a momentum of 0.9 and a weight decay of 0.0005 is employed as the optimizer. The learning rate is initialized to 0.05 and follows a linear warm-up and decay schedule. The entire model is trained end-to-end on Tesla V100 GPU for 40 epochs with a batch size of 16. The number of cached Shape Context Memory features is kept the same as the batchsize during training. For data augmentation, random flipping and rotation are applied to the training data. During inference, if the input size is smaller than the batchsize, we crop the Shape Context Memory features accordingly to match the input size before fusing them in the UNet++ segmentation module.

\begin{table}[htbp]
	\centering
	\renewcommand\arraystretch{1.5}
	\caption{ABLATION STUDY OF THE PROPOSED MODULES ON THE NIH.}
	\label{ablation1}
	\begin{tabular}{ccccc}
		\hline
		% \rowcolor{mygray}
		% MFE  & MAI & HASM & Lasm\mathcal{L}_{asm} & DSC(\%)  \\
		\multicolumn{2}{c}{Baseline} 
		& \multirow{2}*{SCM} 
		& \multirow{2}*{$\mathcal{L}_{asm}$}
		& \multirow{2}*{DSC(\%)}  \\
		\cline{1-2}
		%           % \rowcolor{mygray}
		MFE & MAI&  &  &  \\
		\hline
		\CheckmarkBold &  &  &  & 83.12 \\
		\CheckmarkBold & \CheckmarkBold &  &  & 84.44 \\
		\CheckmarkBold & \CheckmarkBold &  & \CheckmarkBold & 86.14 \\
		\CheckmarkBold & \CheckmarkBold & \CheckmarkBold &  & 89.50 \\
		\CheckmarkBold & \CheckmarkBold & \CheckmarkBold & \CheckmarkBold & 91.00 \\
		\hline
	\end{tabular}
\end{table}

\subsection{Ablation study}
Table~\ref{ablation1} demonstrates the individual contributions of our key components, a) MFE: the multi-scale feature extractor module; b) MAI: the mixed-scale attention integration module; c) Baseline: Only MFE and MAI components are used; d) SCM: the shape context memory module; e) $\mathcal{L}_{asm}$: The ASM prior constraint module.

Our baseline without the proposed SCM and ASM components achieves 84.44\% Dice score on pancreas segmentation. As proposed in Table~\ref{ablation1}, both the SCM and ASM components improve the segmentation performance of the baseline, e.g. SCM boosts Dice by 5\% and ASM boosts by 2\%. Combining SCM and ASM components leads to a 7\% performance gain over the baseline, demonstrating the efficiency and complementarity of the proposed modules. The ablation study thus validates the contributions of our shape context memory module and ASM prior constraint model regularization.
\begin{figure*}[htbp]
	\centerline{\includegraphics[width=0.95\textwidth]{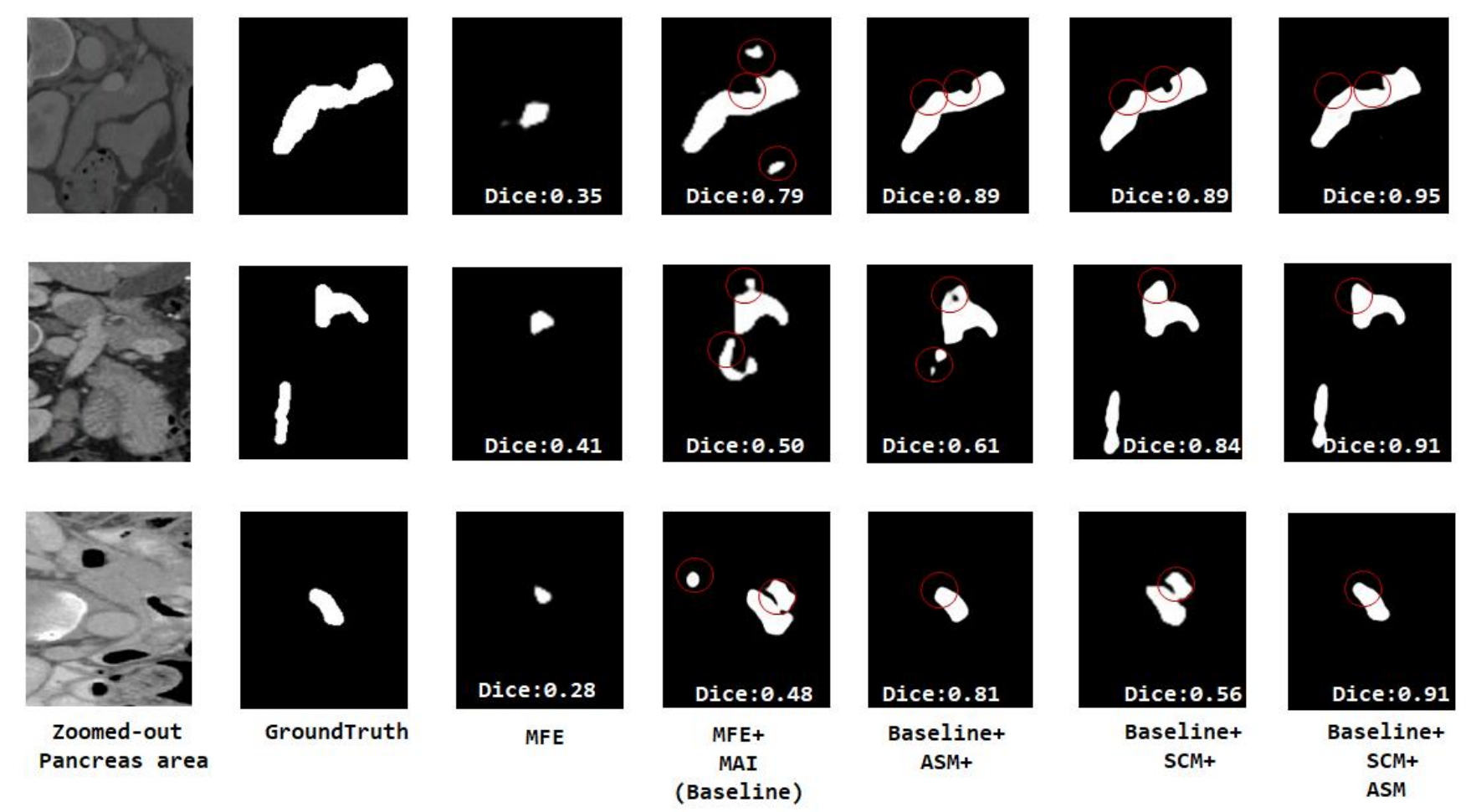}}
	\caption{Visualizes the segmentation improvements from adding different components.}
	\label{fig5}
\end{figure*}

Fig.~\ref{fig5} visualizes the segmentation improvements from adding different components. 
The MAI and MFE modules would reasonably segment regions with indistinct boundaries, ensuring a complete delineation of the pancreatic shape. The ASM prior constraint produces anatomically plausible pancreas shapes. The SCM module effectively captures shape contextual information to implicitly localize the pancreas. ASM and SCM modules complement each other, i.e. the joint model significantly improved segmentation than separate module. Qualitatively, the full model explores the benefits of both shape and context modeling, validating the design of our hybrid architecture. In Fig.~\ref{fig5} the red circles represent locations with significant differences.

\begin{table}[htbp]
	\centering
	\renewcommand\arraystretch{1.5}
	\caption{ABLATION STUDY OF THE PROPOSED SCM MOUDLE ON THE NIH.}
	\label{ablation2}
	\begin{tabular}{cccccc}
		\hline
		% \rowcolor{mygray}
		$F_{1}$  & $F_{2}$ & $F_{3}$ & $F_{4}$ & $F_{5}$ & DSC(\%)  \\
		\hline
		&  & &   \CheckmarkBold &  \CheckmarkBold &   86.61 \\
		&  & \CheckmarkBold &   \CheckmarkBold &  \CheckmarkBold &   87.73 \\
		&  \CheckmarkBold &  \CheckmarkBold&   \CheckmarkBold &  \CheckmarkBold &   89.50 \\
		\CheckmarkBold  & \CheckmarkBold  & \CheckmarkBold &   \CheckmarkBold &  \CheckmarkBold &   88.67 \\
		\hline
	\end{tabular}
\end{table}

\textbf{Shape Context Memory for different Features.} In order to verify which features should be cached in the memory caching to have a better and more positive impact on the final results, we conducted experiments on the cached features in the shape context memory module. The performance of different features for the final segmentation is shown in Table~\ref{ablation2}. The experimental results show that when using the $F_2-F_5$ feature extracted by MAI for shape contextual feature learning, the network shows the best segmentation performance.

% In order to verify which features should be cached in the memory caching module to have a better and more positive impact on the final results, we conducted experiments on the cached features in the hybrid multi-scale attention shape memory-guide module. Firstly, we empirically hypothesized that we would not use the stage 1 features extracted by the HASM module but rather use the stage 2-5 features to be cached, this is because stage 1 feature extraction is too simple and not suitable for our task. To verify this conjecture, we did the following experiments and the results are analyzed in 
Including $F_{i}$ from "$L_1$" of the MAI module degrades performance, as the shallow features lack semantic information.

\begin{table}[htbp]
	\centering
	\renewcommand\arraystretch{1.5}
	\caption{EFFECTIVENESS OF MODELING ASM USING DIFFERENT DATASET.}
	\label{ablationASM1}
	\begin{threeparttable}
		\setlength{\tabcolsep}{7mm}{
			\begin{tabular}{c|ccc}
				\hline
				% \rowcolor{mygray}
				\multirow{2}*{\diagbox[innerwidth=1.5cm]{Train}{Test}}& \multirow{2}*{NIH} & \multirow{2}*{MSD} \\
				\multicolumn{1}{c|}{}& & &  \\
				\hline
				NIH  & 86.14 &  88.32 \\
				MSD  & 85.14  & 89.09 \\
				% MSD+NIH &85.21 & 88.64 \\
				——\tnote{*} & 84.44 & 87.71 \\
				% {\color{blue}{\textbf{English}}}
				\hline
		\end{tabular}}
		\begin{tablenotes}    % 添加命令
			\footnotesize               % 添加命令
			\item[*] "——" denotes baseline network without $\mathcal{L}_{asm}$ .		  %自己改注释 和表格对应
		\end{tablenotes}  
		% 添加命令
	\end{threeparttable}       % 添加命令
	
\end{table}
\textbf{ASM learning using different Dataset.} 
%We validate the proposed ASM prior constraint module via ablation studies in Table~\ref{ablationASM1}. ASM prior constraint models are built using the NIH dataset, MSD dataset, or their combination, and evaluated on both datasets. The results show ASM contributes to improved performance, however MSD underperforms NIH for cross-dataset evaluation due to MSD containing atypical lesion-induced shape changes. Surprisingly, combining datasets is ineffective compared to NIH alone, despite providing more shape diversity. Given the generality of our ASM formulation, we select the NIH-based model as it transfers better across datasets.
%We validated the effect of ASM models trained on different datasets on network segmentation performance using the ablation studies in Table~\ref{ablationASM1}. The ASM models were trained using NIH, MSD, or a combination of both, and evaluated on both datasets. The results show that the trained ASM models $L_{asm}$ both improve over the baseline segmentation. The NIH-based ASM model achieves optimal performance on NIH and competitive results on MSD. MSD contains shape variations caused by atypical lesions, so the MSD-based ASM model is less generalizable. Surprisingly, the combined dataset underperforms the dataset using NIH alone, despite the increased shape diversity. Given the generalizability of our formulation, the ASM model in our $L_{asm}$ was ultimately chosen to be trained on the NIH dataset.
We verified the effect of ASM models trained on different datasets, in terms of segmentation performanc. As shown in Table~\ref{ablationASM1}, when employing $\mathcal{L}_{asm}$, the network achieves better segmentation performance than baseline network on both datasets, which demonstrates the effectiveness of $\mathcal{L}_{asm}$. It is noticed that the ASM prior constraint module shows good cross-dataset performance. The module trained using NIH and MSD dataset achieve 0.6\% and 0.7\% higher Dice on MSD and NIH dataset, respectively, than baseline network. Since the MSD contains shape variations caused by atypical lesions, and thus the MSD-based ASM model has lower cross-dataset modeling capability. As a result, we finally choice to train the ASM model for $\mathcal{L}_{asm}$ using the NIH dataset.
%showed that the ASM models performed better than the baseline under different data performances, demonstrating that the ASM models have good cross-dataset modeling capabilities.The ASM models were trained using either the NIH or the MSD, and evaluated on both datasets. The NIH-based ASM model achieved the best performance on the NIH dataset and competitive results on the MSD dataset.The MSD contains shape variations caused by atypical lesions, and thus the MSD-based ASM model has poor cross-dataset modeling ability. Considering the generalization ability of our proposed model, the final choice is to train the ASM model in $\mathcal{L}_{asm}$ on the NIH dataset.
\begin{table}[htbp]
	\centering
	\renewcommand\arraystretch{1.5}
	\caption{COMPARE STATE-OF-THE-ART METHODS ON THE NIH.}
	\label{stateNIH}
		\resizebox{\columnwidth}{!}{%
	\begin{threeparttable}
		\begin{tabular}{ccccc}
			\hline
			% \rowcolor{mygray}
			\multicolumn{2}{c}{Methods}
			%Methods & DSC(\%)\tnote{*}  & Precision(\%)\tnote{*}  & Recall(\%)\tnote{*}   \\
			&
			DSC(\%)\tnote{*}  & Precision(\%)\tnote{*}  & Recall(\%)\tnote{*} \\
			\hline
			\multirow{6}{*}{CNN-Based}
			&
			M. Li et al \cite{n38}. & 83.31 $\pm$ 6.32 & 84.09 $\pm$ 8.65 & 83.30 $\pm$ 8.54 \\
			&
			Y. Zhang et al \cite{n39}. & 84.47 $\pm$ 4.36 & - & -\\
			&
			D. Zhang et al \cite{n15}. & 84.90 & - & -  \\
			&
			H. Chen et al \cite{n40}. & 85.19 $\pm$ 4.73 & 86.09 $\pm$ 5.93 & 84.58 $\pm$ 8.09  \\
			&
			J. Li et al \cite{n33}. & 83.35 $\pm$ 4.13 & 83.45 $\pm$ 7.19 & 82.76 $\pm$ 8.21 \\
			&
			p. Hu et al \cite{n41}. & 85.49 $\pm$ 4.77 & - & - \\
			&
			H. Li et al \cite{n42}. & 85.70 $\pm$ 4.10 & 87.40 $\pm$ 5.20 & 84.80 $\pm$ 7.50  \\
			&
			F. Li et al \cite{n43}. & 87.57 $\pm$ 3.26 & 86.63 $\pm$ 3.70 & 89.55 $\pm$ 4.03  \\
				&
			\textcolor{red}{\textbf{Ours}} & \textcolor{red}{\textbf{91.00 $\pm$ 1.33}} & \textcolor{red}{\textbf{91.26 $\pm$ 2.83}} & \textcolor{red}{\textbf{91.89 $\pm$ 2.77}}  \\
			\hline
			\multirow{2}{*}{Transform-Based} 
			&
			C. Qui et al \cite{n57}. & 86.25 $\pm$ 4.52 & - & -  \\
			&
			S. Dai et al \cite{n44}. & 89.89 $\pm$ 1.82 & 89.59 $\pm$ 1.75 & 91.13 $\pm$ 1.48  \\
			% {\color{blue}{\textbf{English}}}
			% \rowcolor{mygray}
			\hline
		\end{tabular}
		\begin{tablenotes}    % 添加命令
			\footnotesize               % 添加命令
			\item[*] The results (measured by the DSC, Precision, Recall) of pancreas segmentation on the NIH dataset. “—” denotes that the corresponding results are not provided in the literature. Optimal results (described by mean ± std) are shown in red bold.		  %自己改注释 和表格对应
		\end{tablenotes}            % 添加命令
	\end{threeparttable}  }     % 添加命令
	
\end{table}

\subsection{Comparison with state-of-the-art methods}\label{comparsion}
\textbf{NIH dataset.} We compare our model against state-of-the-art methods \cite{n15,n33,n38,n39,n40,n41,n42,n43,n44} on the NIH dataset using four-fold cross-validation. As shown in Table~\ref{stateNIH}, our proposed model achieves the state-of-the-art Dice score of 91.00\%, outperforming prior arts. The low standard deviation of 1.33 demonstrates robustness across different cases. Our model also maintains high precision and recall, highlighting its potential. The superior accuracy and robustness validate the effectiveness of our shape context memory module and ASM prior constraint module.

\begin{table}[htbp]
	\centering
	\renewcommand\arraystretch{1.5}
	\caption{COMPARE STATE-OF-THE-ART METHODS ON THE MSD.}
	\label{stateMSD}
	\resizebox{\columnwidth}{!}{%
	\begin{threeparttable}
		
		\begin{tabular}{ccccc}
			\hline
			% \rowcolor{mygray}
			\multicolumn{2}{c}{Methods}
			%Methods & /
			&
			DSC(\%)\tnote{*}  & Precision(\%)\tnote{*}  & Recall(\%)\tnote{*}   \\
			\hline
			\multirow{6}{*}{CNN-Based}
			&
			H. Chen et al \cite{n40}. & 76.60 $\pm$ 7.30 & 87.70 $\pm$ 8.30 & 69.20 $\pm$ 12.80 \\
			&
			Y. Zhang et al \cite{n39}. & 82.74  & - & -\\
			&
			D. Zhang et al \cite{n15}. & 85.56 & - & -  \\
			&
			J. Li et al \cite{n33}. & 85.65  & - & -  \\
			&
			W. Li et al \cite{n17}. & 88.52 $\pm$ 3.77 & - & 91.86 $\pm$ 5.06 \\
			&
			\textcolor{red}{\textbf{Ours}} & \textcolor{red}{\textbf{92.25 $\pm$ 1.12}} & \textcolor{red}{\textbf{94.79 $\pm$ 3.25}} & \textcolor{red}{\textbf{92.77 $\pm$ 1.88}}  \\
			\hline
			\multirow{1}{*}{Transform-Based} 
			&
			S. Dai et al \cite{n44}. & 91.22 $\pm$ 1.37 & 93.22 $\pm$ 2.79 & 91.35 $\pm$ 1.63  \\
			% {\color{blue}{\textbf{English}}}
			% \rowcolor{mygray}
			\hline
		\end{tabular}
		\begin{tablenotes}    % 添加命令
			\footnotesize               % 添加命令
			\item[*] The results (measured by the DSC, Precision, Recall) of pancreas segmentation on the MSD dataset. “—” denotes that the corresponding results are not provided in the literature. Optimal results (described by mean ± std) are shown in red bold.		  %自己改注释 和表格对应
		\end{tablenotes}            % 添加命令
	\end{threeparttable}   }    % 添加命令
	
\end{table}

\textbf{MSD dataset.} 
Table \ref{stateMSD} shows the comparison of the proposed network with SOTA methods \cite{n15,n17,n33,n39,n40,n44} on the MSD. From Table \ref{stateMSD} we can see that our network achieves a superior DSC of 92.25 \%. The state-of-the-art results demonstrate the robustness of our proposed network.

\section{CONCLUSION}
To address the challenges of boundary ambiguity and large shape variations in pancreas segmentation, we propose a novel framework that consists of multi-scale Feature Extraction Module (MFE), mixed-scale attention integration module (MAI), shape context memory module (SCM), and a ASM prior constraint module, which combines the idea of multi-scale feature extraction and anatomical shape prior. Specifically, MFE and MAI are employed to solve the edge blur problem of the pancreas and obtain a reasonable pancreas region; the SCM implicitly encodes spatial and shape cues, in conjunction with the ASM prior constraint module, to guide shape-aware segmentation. Extensive experiments demonstrate the state-of-the-art performance and validate the efficiency of the SCM and ASM prior constraint components through ablation studies. Furthermore, our framework provides a generalizable approach for other medical image segmentation tasks. In future work, we will further investigate its application to other organ segmentation problems.
\section*{ACKNOWLEDGMENT}
This work was supported by the National Natural Science Foundation of China under Grant 82261138629; Guangdong Basic and Applied Basic Research Foundation under Grant 2023A1515010688 and 2021A1515220072; Shenzhen Municipal Science and Technology Innovation Council under Grant JCYJ20220531101412030 and JCYJ20220530155811025.
% \printbibliography{}
\bibliographystyle{IEEEtran}
\bibliography{SCPMan}
% \end{thebibliography}

\end{document}